\documentclass[letterpaper, 10 pt, conference]{IEEEtran}

\usepackage{graphicx} 
\usepackage{xcolor} 
\usepackage{multirow}
\usepackage{tabularray}
\usepackage{booktabs}
\usepackage{pgfplots}
\usepackage{cite}
\usepackage{array}
\usepackage{amsmath} 
\usepackage[table]{xcolor}

\usepackage[font=footnotesize,labelsep=period]{caption}
\usepackage{balance}
\begin{document}

\title{Evaluation of Vision-Language Models Across Diverse Coastal Environments}
\author{Seth Knoop, Chad R. Samuelson, Gabriel R. Slade, Brady Moon, and Joshua G. Mangelson
    \thanks{This work was partially funded under Office of Naval Research award numbers N00014-24-1-2301.}%
  \thanks{S.~Knoop, C.~Samuelson, G.~Slade, B.~Moon, and J.~Mangelson are at Brigham Young University. They can be reached at: \texttt{\{sk8723, chadrs2, grs45, brady.moon, mangelson\}@byu.edu}. }
}
\maketitle




\begin{abstract}
Vision-language models (VLMs) enable robotic perception by associating visual observations with natural-language concepts.
Yet their performance in coastal environments remains largely unexplored. 
We introduce a densely labeled coastal dataset containing more than 1,000 images collected across seven missions in three regions of Oahu, Hawaii, with 18 semantic classes and over 7,400 annotated instances. 
We evaluate seven modern VLMs through three complementary experiments measuring text-to-mask, mask-to-mask, and mask-to-text alignment. 
Broad landscape classes are generally recognized more accurately than conventional object and coastal classes, with coastal concepts presenting the greatest challenge. 
However, comparisons of shared conventional classes across coastal and terrestrial datasets reveal no consistent performance difference attributable solely to environmental context. 
Mask-to-mask matching also remains similar across conventional and coastal classes, while alternative textual labels substantially improve recognition of several coastal concepts. 
These results suggest that lower performance on coastal classes (at least on the objects/query categories evaluated) is heavily influenced by segmentation and linguistic representation.
\end{abstract}

\section{Introduction}
\label{sec:intro}

Vision-Language Models (VLMs) have emerged as useful tools for robotic systems to better understand and reason about the world around them through visual and textual modalities.
Existing computer vision benchmark datasets are built predominantly for terrestrial and indoor spaces, limiting VLM evaluation to a particular range of environmental domains and features.
As a result, VLM performance in littoral domains remains largely unexplored, limiting standardized evaluation and comparability across studies.

Coastal environments exhibit features absent or rare in land-based environments.
Ambiguous shoreline boundaries weaken training signal reliability and under-represented marine object classes limit robust mapping between visual inputs and semantic labels.
These limitations directly impact autonomous marine robotic systems that rely on VLMs to perform perception-based tasks using natural language, degrading risk assessment and navigation decisions, potentially leading to unsafe real-world deployments.

In this work, we evaluate several state-of-the-art (SOTA) VLMs in coastal environments to assess their ability to generalize beyond conventional benchmark datasets.
These evaluations are designed to expose the intra- and inter-modality capabilities and gaps of VLMs in coastal environments.

\begin{figure}[t]
    \centering
    \includegraphics[width=\columnwidth]{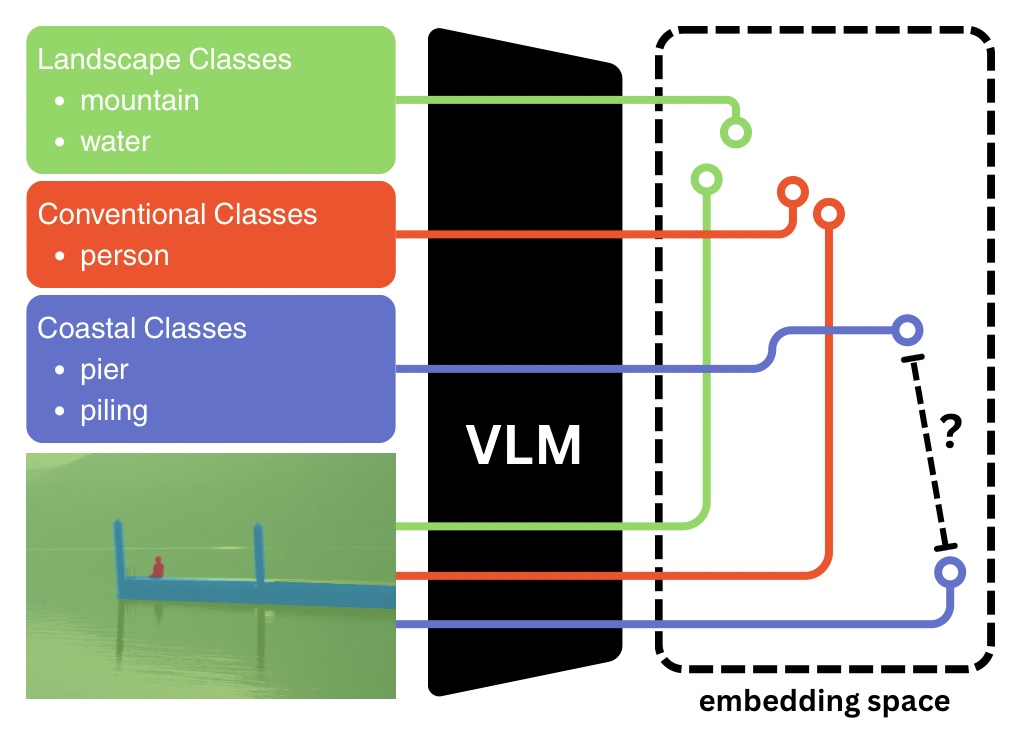}
    \caption{There is a gap in open-set VLM semantic embedding of unique marine or coastal classes.}
    \label{fig:cover_fig}
\end{figure}

\section{Related Work}
\label{sec:related_work}

\subsection{Vision-Language Representation Learning}

VLMs learn the relationships between visual features and natural-language descriptions, enabling visual concepts to be represented in a language-aligned space.
CLIP~\cite{clip_radfordLearningTransferable2021} demonstrated that large-scale pretraining on image-text pairs can produce representations that enable zero-shot recognition using natural-language labels.
Subsequent approaches have explored alternative training methods and architectures, including SigLIP~\cite{zhai2023siglip}, to increase quality and scalability.
These approaches provide a scalable alternative to conventional closed-set recognition, in which the set of semantic categories must be formally defined during training, given that the training distribution accurately reflects the deployment distribution. 

However, image-level vision-language alignment does not necessarily imply that semantic information extends to the level of individual objects or regions.
This has motivated methods that expand vision-language representations to dense visual features and spatially localized predictions.

\subsection{Applications of Vision-Language Alignment }

Dense self-supervised visual representations, such as DINO~\cite{simeoni2025dinov3}, have demonstrated that pretrained vision transformers can encode semantically meaningful local visual features.
Grounding-DINO~\cite{gdino2024} extends these visual representations through language-grounded pretraining for open-set object detection, enabling recognition beyond the limitations of conventional closed-set approaches.
Similarly, CLIP-DINOiser~\cite{clipdinoiser2024} leverages the rich visual features of DINO to guide the learning of dense CLIP features from MaskCLIP~\cite{zhou2022maskclip}, enabling object segmentation given natural-language prompts. 
The Segment Anything Model (SAM~\cite{kirillov2023sam}), by contrast, provides class-agnostic segmentation capabilities, producing object masks without predefined semantic categories.
Grounded-SAM~\cite{ren2024gsam} combines the language-grounded localization of Grounding-DINO with the segmentation capabilities of SAM, associating object masks with natural-language concepts.
Rather than combining separately trained models at inference time, RADIO~\cite{ranzinger2024radio} distills multiple vision foundation models (VFMs), including CLIP, DINO, and SAM, into a single backbone that produces dense features aligned with visual, language, and segmentation representations.

Together, these approaches support complementary forms of vision-language understanding, including visual grounding, open-vocabulary segmentation, and cross-modal alignment.
In this work, we evaluate these complementary capabilities through a series of experiments.

\subsection{Coastal Domain Gap}

Many influential computer vision benchmarks are dominated by terrestrial imagery and conventional object categories.
For example, COCO~\cite{lin2015microsoft} provides a large-scale benchmark containing more than 100,000 images and a broad set of object and semantic categories.
Recently, datasets such as GOOSE~\cite{mortimer2024goosedataset} provide additional coverage of complex terrestrial environments.
These datasets have been integral in developing and validating modern VLMs.
However, they provide comparatively limited coverage of coastal environments and maritime categories.

Existing maritime datasets are often designed around specific predefined categories.
For example, the LaRS~\cite{zust2023lars} dataset provides annotated imagery for surface vehicle perception, focusing on boats and waterborne people.
However, categories representing key maritime infrastructure--such as piers and wharves--are not included.
Consequently, existing benchmarks provide limited opportunities to evaluate whether vision-language representations can accurately recognize coastal concepts.
This underrepresentation creates a meaningful domain gap for evaluating the generalization of VLMs to coastal environments.

\section{Coastal Dataset}
\label{sec:dataset}

\begin{figure}[t]
    \centering
    \includegraphics[width=\columnwidth]{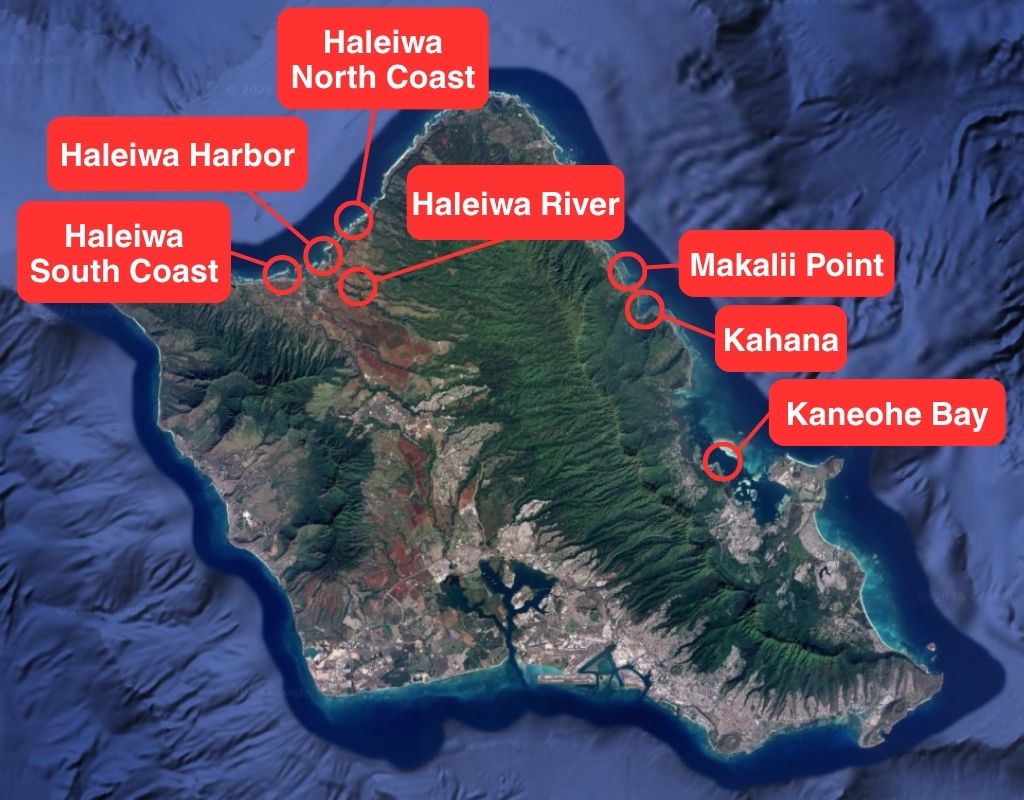}
    \caption{Map of locations where data was collected.}
    \label{fig:data_collection_map}
\end{figure}

To evaluate VLMs in coastal environments, we leverage a highly diverse labeled image dataset collected across seven missions in Oahu, Hawaii (see Fig.~\ref{fig:data_collection_map}). 
The dataset spans three geographically distinct regions, including harbors, rivers, and open coastal waters.

\begin{figure}[t]
    \centering
    \includegraphics[width=\columnwidth]{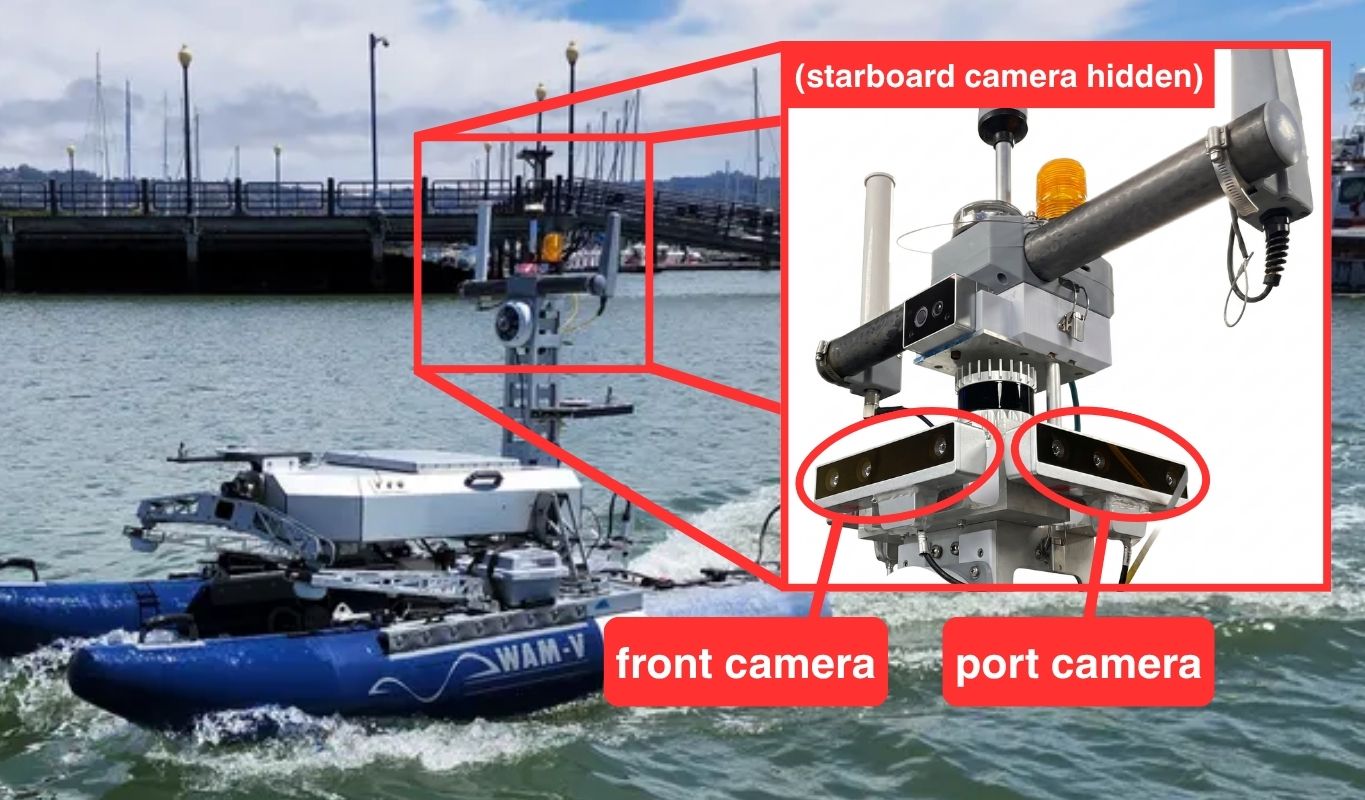}
    \caption{WAM-V custom sensor platform featuring three OAK-D Long Range stereo cameras surrounding its base and an OS1 LiDAR on top.
    }
    \label{fig:sensor_platform}
\end{figure}

Data were collected from a Wave Adaptive Modular Vessel (WAM-V) equipped with a custom sensor platform featuring three OAK-D Long Range stereo cameras oriented forward, port, and starboard (see Fig.~\ref{fig:sensor_platform}).
Missions were ran with the WAM-V at a variety of different angles and distances from the shoreline.
The raw sensor data was originally recorded as ROS bag files, from which RGB images were extracted and downsampled into a curated set of 1,000+ images to promote labeling efficiency while maintaining feature variance.

Each image was manually labeled using Roboflow~\cite{roboflow2024} to produce per-pixel masks that support both semantic and instance-level segmentation. 
We selected classes according to three criteria: (1) \textbf{prevalence}, the extent to which the class was represented across the dataset ensuring sufficient instances for meaningful analysis, (2) \textbf{visual distinctiveness}, the degree to which the class could be consistently identified by human annotators, and (3) \textbf{scientific relevance}, the extent to which the class either represented a unique feature in coastal environments or corresponded to an existing category in another established dataset.

\begin{figure}[t]
    \centering
    \includegraphics[trim={0cm 0.6cm 0cm 0.8cm},clip,width=\columnwidth]{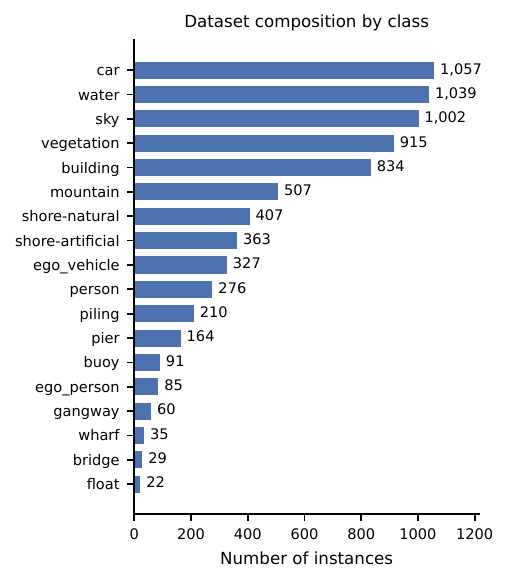}
    \caption{Dataset composition by class.}
    \label{fig:dataset_composition}
\end{figure}

The dataset comprises 18 classes and more than 7,400 annotated instances (see Fig.~\ref{fig:dataset_composition}).
While several classes are visually straightforward, others require more specific definitions. There are two shore classes: \texttt{shore-natural} defined as naturally formed terrain occurring between water and mainland, including beaches and banks, and \texttt{shore-artificial}, defined as engineered shoreline structures that form a transition between land and water, such as revetments and breakwaters.
Scene-level classes were annotated as a single instance per class within each image.
These classes include \texttt{mountain}, \texttt{sky}, \texttt{vegetation}, \texttt{shore-natural}, \texttt{shore-artificial}, and \texttt{water}, whereas discrete objects were annotated by instance. 

The dataset also contains several maritime structures.
A \texttt{pier} is defined as a structure extending approximately perpendicular from the mainland into the water, whereas a \texttt{wharf} is a structure that is built parallel to the mainland that provides access for people and waterborne vessels.
Finally, a \texttt{float} is defined as a floating structure that is completely surrounded by water.

Two classes, referred to as ``void classes", are excluded from evaluation to prevent over-representation biases in model evaluation and are subtracted from other overlapping annotations: \texttt{ego\_vehicle} corresponds to a portion of the WAM-V that occasionally appears within the camera frame, while \texttt{ego\_person} represents members of the research team captured in the images.

\section{Evaluation Methodology}
\label{sec:method}

Our evaluation compares different aspects of modern VLM capabilities across coastal and terrestrial datasets to identify where performance differs between these domains.
To characterize these differences, we conduct three experiments designed to probe intra- and inter-modal capabilities across seven modern VLMs. 

The first and third experiments evaluate text-to-mask and mask-to-text performance, respectively, testing each model's ability to ground semantic concepts in visual observations and, conversely, associate visual observations with semantic concepts.
These experiments evaluate inter-modal capabilities.
The second experiment evaluates intra-modal semantic alignment by measuring how effectively models associate semantically similar visual regions across observations through mask-to-mask matching.

\subsection{Models}
\label{subsec:models}

To evaluate the generalization capability of current SOTA VLMs, we consider a diverse set of architectures including CLIP~\cite{clip_radfordLearningTransferable2021}, CLIP-DINOiser~\cite{clipdinoiser2024}, Grounded-SAM~\cite{ren2024gsam}, C-RADIOv3~\cite{ranzinger2024radio}, SAM3~\cite{carion2025sam3segmentconcepts}, SigLIP~\cite{zhai2023siglip}, and YOLOE-seg~\cite{wang2025yoloerealtimeseeing}.
For models that do not produce segmentation masks directly, CLIP and SigLIP, we use FastSAM~\cite{zhao2023fastsam} as a shared mask-generation module.

To facilitate fair comparisons, we selected VLM variants with comparable parameter counts where possible. 
Specifically, we evaluated ViT-B/32 (CLIP), ViT-B/16 (CLIP-DINOiser), SwinT+ViT-B (Grounded-SAM), Base+SigLIP2 (C-RADIOv3), and ViT-B/16 (SigLIP), which have approximately 150--300M parameters. 
Since YOLOE-seg's largest model is around 70M parameters, we selected the middle model, YOLOE-m-seg, to maintain consistency with the base-sized variants used for the other models.
SAM3 is an outlier, as only one model variant is available, with approximately 850M parameters.

All models are evaluated without domain-specific tuning to isolate out-of-the-box generalization performance.
However, as some CLIP-style models are sensitive to prompt formation, we adopt the prompt templates specified in the respective original works to ensure that each model is evaluated with its intended prompting strategy and maintain fair comparison.
CLIP uses ``\texttt{a photo of a <class>}" as proposed by the original paper \cite{clip_radfordLearningTransferable2021}, and so does SigLIP~\cite{zhai2023siglip}.
CLIP-DINOiser~\cite{clipdinoiser2024} includes a \texttt{background} prompt to handle all image portions that don't match the natural language prompts.
Grounded-SAM represents multiple classes by separating them with periods, using the format ``\texttt{<class1> . <class2> . <class3>}" as in \cite{gdino2024}.

\subsection{Experiment 1: Input Text; Output Semantic Mask}
\label{subsec:exp1_text2mask}

The first experiment attempts to answer the question: \textit{How well do VLMs understand semantic concepts and localize them correctly in coastal visual observations?}
In this experiment, we provide VLMs with a set of natural language class names of the format ``\texttt{<class>}" and a set of RGB images.
The VLMs then predict regions within each image that align with the provided natural language prompts.
Performance metrics are then calculated pixel-wise on how well these predicted regions coincide with the ground truth annotations.

Grounded-SAM, SAM3, and YOLOE-seg models automatically output instance segmentations provided a list of natural language prompts.

Since CLIP and SigLIP do not directly generate segmentation masks, we follow a common approach that first generates class-agnostic masks and then assigns them to natural language prompts \cite{maggio2024clio, samuelson2025terra}. 
We use FastSAM to generate approximately 100 masks per image. 
Each mask is passed through CLIP or SigLIP to produce a visual embedding, while the prompted class names are used to generate corresponding text embeddings. 
We then compare mask and text embeddings using cosine-similarity. 
Masks with similarities above $0.25$ for CLIP and $0.07$ for SigLIP are assigned to the corresponding text prompt. 
These thresholds were selected based on previous work~\cite{samuelson2025terra} and preliminary experiments that achieved the highest mean F1-scores across classes in the Hawaii dataset.

C-RADIOv3 outputs a set of image patch embeddings.
Using a SigLIP-2 adaptor, we generate a set of natural language embeddings that are aligned in the RADIO embedding space.
We then compute cosine-similarity between the image patch embeddings and each text prompt embedding.
Patches with similarities above $0.07$ are assigned to the corresponding text prompt. 
The same threshold-selection procedure used for CLIP and SigLIP was applied to C-RADIOv3, emphasizing the highest mean F1-score across classes.

The metrics used are mean intersection over union (mIoU), mean F1-score (mF1), and mean average precision (mAP).
For pixelwise segmentation metrics, we use the standard definitions of true positives (TP), false positives (FP), false negatives (FN), and true negatives (TN) with regards to pixel level accuracy.
Let $C$ denote the total number of text prompts, the performance metrics are then defined as:
\begin{equation}
    \text{mIoU} = \frac{1}{C} \sum^C_{c=1} \frac{TP_c}{TP_c + FN_c + FP_c},
\end{equation}
\begin{equation}
    \text{mF1} = \frac{1}{C} \sum^C_{c=1} \frac{2 \cdot TP_c}{2 \cdot TP_c + FN_c + FP_c},
\end{equation}
\begin{equation}
    \text{mAP}_{50} = \frac{1}{C} \sum^C_{c=1} AP_c(0.50),
\end{equation}
where predictions with $\text{IoU} \geq 0.50$ are considered true positive.
$\text{mAP}_{75}$ is similarly defined as $\text{mAP}_{50}$ but with $0.75$ as the threshold.
Lastly, the general $\text{mAP}$ metric is defined as:
\begin{equation}
    \text{mAP}_{50:95} = \frac{1}{10 \cdot C} \sum_{c=1}^C \sum_{\tau \in \{ 0.50, 0.55, \cdots, 0.95 \} } AP_c(\tau).
\end{equation}

Since CLIP-DINOiser and C-RADIOv3 produce dense patch-level features rather than instance-level masks, we don't report mAP performance metrics for these models.

\subsection{Experiment 2: Input Mask; Output Top-K Semantic Masks}
\label{subsec:exp2_mask2mask}

The second experiment asks: \textit{When given a ground-truth region mask, how well can a given VLM identify other masks with the same semantic class?}
We provide each VLM with the ground-truth masks from the labeled dataset and generate an embedding for each mask. 
For each mask, we compute its cosine-similarity with every other mask in the dataset and rank the masks by similarity.
We then select the top-$k$ most similar masks.
An association is considered correct if at least one of the top-$k$ masks belongs to the same class as the query mask, yielding an accuracy of $1.0$.

CLIP and SigLIP encode each mask directly as a single embedding, while CLIP-DINOiser, Grounded-SAM, SAM3, and C-RADIOv3 use masked average pooling over dense visual features. 
CLIP-DINOiser uses MaskCLIP features refined with DINO attention maps, Grounded-SAM pools and concatenates multi-scale Grounding-DINO features, and C-RADIOv3 and SAM3 pool features from their respective vision backbones. 
The resulting embeddings are compared using cosine similarity and ranked for top-$k$ evaluation.
YOLOE-seg is excluded from this experiment, as its visual encoder features are not accessible.


The metrics used in this experiment are top-$k$ accuracy and excess over chance.
For each query mask, top-$k$ accuracy measures whether at least one of the $k$ most similar masks belongs to the same class as the query mask.
Let $N$ denote the total number of mask instances, and let $I_i^{(k)}$ be an indicator that is $1$ if at least one of the top-$k$ retrieved masks for query mask $i$ belongs to the same class, and $0$ otherwise.
Top-$k$ accuracy is defined as:
\begin{equation}
    \text{Top-$k$ Accuracy} = \frac{1}{N}\sum_{i=1}^{N} I_i^{(k)}.
\end{equation}


Classes containing fewer than two instances are excluded from this experiment because a query from such a class has no other same-class instance that can be retrieved.

\subsection{Experiment 3: Input Mask; Output Top-K Text Classes}
\label{subsec:exp3_mask2text}

The third experiment asks: \textit{Given a ground-truth region mask, how well can a given VLM classify its semantic class from a selection of words associated with each class?}
We provide each VLM with the ground-truth masks from the labeled dataset and generate an embedding for each mask. 
We also provide each VLM with a word bank and generate an embedding for each word.
For each mask embedding, we compute its cosine-similarity with every word embedding in the word bank and select the top-$k$ most similar words.
An association is considered correct if at least one of the top-$k$ words represents the same class as the query mask, yielding an accuracy of $1.0$.

Each model generates a visual embedding for each ground-truth mask following the procedures described in Experiment 2 (Section~\ref{subsec:exp2_mask2mask}) and then generates a text embedding for each word in the word bank.
Each approach ensures that the visual and text embeddings share the same embedding space so cosine-similarity is computed directly between them

The models differ in how they generate text embeddings.
CLIP and SigLIP directly encode each word in the word bank using their respective text encoders.
C-RADIOv3 utilizes its SigLIP2 adaptor's text encoder.
CLIP-DINOiser uses CLIP's text encoder.

Grounded-SAM and SAM3 do not produce text embeddings and visual embeddings in a shared embedding space and are therefore excluded from this experiment as their embeddings cannot be directly compared.
Similar to Experiment 2, YOLOE-seg is excluded from this experiment, as its visual encoder features are not accessible.

This experiment uses top-$k$ accuracy as described in Section~\ref{subsec:exp2_mask2mask}.

\section{Results}
\label{sec:results}

\subsection{Experiment 1: Text-to-Semantic Mask}
\label{subsec:results_exp1}

Experiment 1 evaluates the ability of each model to associate natural-language class descriptions with the corresponding regions in RGB images.
Table~\ref{tab:exp1_overall} summarizes the performance of each model on the Hawaii dataset using metrics defined in Section~\ref{subsec:exp1_text2mask}.

\begin{table}[t]
\centering
\begin{tabular}{l|lllll}
    \hline
    Model & $\text{IoU}$& $\text{mF1}$& $\text{mAP}$& $\text{mAP}_{50}$& $\text{mAP}_{75}$\\
    \hline
    CLIP & 0.1005& 0.1446& 0.0195& 0.0334& 0.0199\\
    CLIP-DINOiser & 0.1208& 0.1938& -& -& -\\
    Grounded-SAM & 0.3279& 0.4284& 0.1585& 0.2412& 0.1720\\
    C-RADIOv3 & 0.2150& 0.2809& -& -& -\\
    SAM3 & \textbf{0.4023}& \textbf{0.5069}& \textbf{0.3425}& \textbf{0.4854}& \textbf{0.3760}\\
    SigLIP & 0.1456& 0.2149& 0.0361& 0.0771& 0.0308\\
    YOLOE-seg & 0.1451& 0.2206& 0.0723& 0.1427& 0.0609\\
    \hline
\end{tabular}
\caption{Text-to-mask Performance on the Hawaii dataset. ``-" indicates that the metric is inapplicable to the given VLM.}
\label{tab:exp1_overall}
\end{table}

Table~\ref{tab:exp1_overall} shows that SAM3 achieves the highest mIoU, mF1, and mAP among the evaluated models.
Its advantage may partly reflect its substantially larger model capacity, although this experiment does not isolate model size from architectural and training differences. 
Thus, the results should be interpreted as overall performance differences rather than evidence of a direct relationship between model size and performance.
CLIP and SigLIP do not natively generate masks and therefore use FastSAM-proposed masks. 
Their performance reflects both the quality of the proposed masks and vision-language alignment, rather than semantic alignment alone. 
Additionally, similarity thresholds for CLIP, SigLIP, and C-RADIOv3 were selected by maximizing mF1 on the Hawaii dataset. 
Although the model weights were not tuned on this dataset, this threshold selection introduces dataset-specific information into the evaluation. 
We therefore characterize these results as zero-shot with respect to model weights, but not fully zero-shot at the pipeline level.
Each proceeding result should be understood in the context of these limitations.

\begin{figure}[t]
    \centering
    \includegraphics[trim={0cm 0.3cm 0cm 0.0cm},clip,width=\columnwidth]{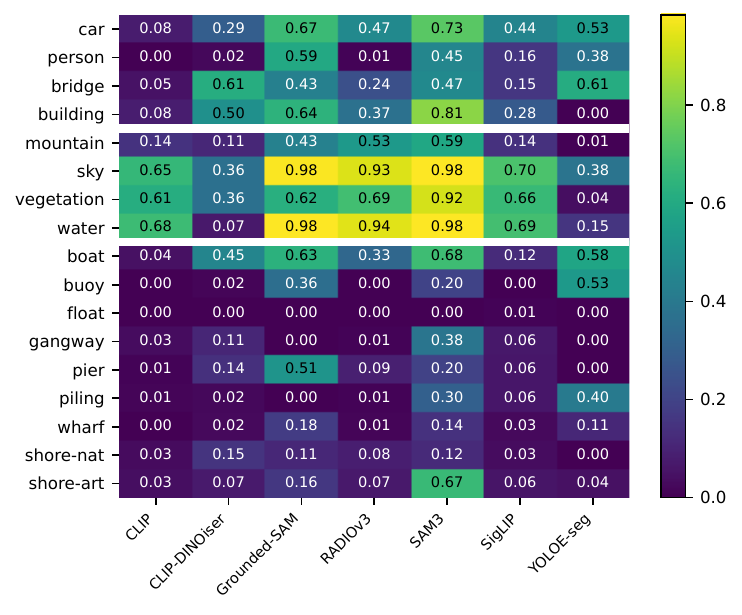}
    \caption{Text-to-mask F1-scores of each class in the Hawaii dataset.}
    \label{fig:exp1_perclass_heatmap_F1}
\end{figure}

Beyond overall performance, class-level evaluation provides greater insight into the challenges of segmentation.
Fig.~\ref{fig:exp1_perclass_heatmap_F1} shows the F1-score for each class and model in the Hawaii dataset.
The classes shown in the heat map are divided into three semantic groups in the figure: conventional (top four), landscape (middle four), and coastal (bottom nine).
A clear pattern in Fig.~\ref{fig:exp1_perclass_heatmap_F1} is the difference in performance between the three semantic groups.
The landscape classes generally achieve the highest F1-scores, followed by the conventional classes, and lastly the coastal classes with the lowest F1-scores.

However, this difference may reflect both semantic specificity and the greater segmentation difficulty associated with classes that are small.
To characterize this factor, we compute Spearman's rank correlation~\cite{spearman1904}, with the results reported in Table~\ref{tab:exp1_spearman_corr}.
Spearman's rank correlation is calculated between per-class F1-score and the average class instance size per image in the Hawaii dataset.
Since classes in the landscape group generally occupy a large proportion of the image and have higher F1-scores, including the landscape classes substantially biased the observed correlation values. 
We therefore excluded them to focus the analysis on the relationship between conventional and coastal classes.

The correlation between spatial extent and segmentation performance is weak and inconsistent across models.
The $\rho$ values vary in sign and magnitude, ranging from $-0.2253$ to $0.3846$, and none reach a statistical significance of $p<0.05$.
This provides limited evidence that the proportion of an image occupied by a class is associated with segmentation performance.
This leaves open the possibility that characteristics specific to coastal classes or environmental context contribute to their weaker segmentation performance.

\begin{table}[t]
\centering
\begin{tabular}{lll}
    \hline
    Model & $\rho$& $p$\\
    \hline
    CLIP & 0.1648 & 0.5905 \\
    CLIP-DINOiser & 0.3846 & 0.1944 \\
    Grounded-SAM & -0.1823 & 0.5511 \\
    C-RADIOv3 & 0.3297 & 0.2713 \\
    SAM3 & -0.1538 & 0.6158 \\
    SigLIP & -0.2253 & 0.4593 \\
    YOLOE-seg & -0.1074 & 0.7269 \\
    \hline
\end{tabular}
\caption{Spearman's rank correlation between per-class F1-score and class level characteristics in the Hawaii dataset.}
\label{tab:exp1_spearman_corr}
\end{table}

To evaluate the effects of environmental context, Table~\ref{tab:exp1_conclass_multidataset_F1} compares the F1-scores of the same conventional classes across three datasets (Hawaii, COCO, and GOOSE).
For consistency, we use a shared set of conventional classes that are present in all three datasets--namely \texttt{car}, \texttt{person}, \texttt{bridge}, and \texttt{building}.
The results show no consistent trend in cross-dataset performance.
While some models exhibit higher mF1 scores on Hawaii than on COCO or GOOSE, others perform worse, with the magnitude and direction varying significantly across models.
The observed differences do not reveal a consistent relationship between coastal and terrestrial context.
Further analysis would therefore be required to explore the potential affect of domain shift to account for the varied performance.

\begin{table}[t]
\centering
\begin{tabular}{llll}
    \hline
    Model & Hawaii & COCO & GOOSE \\
    \hline
    CLIP & 0.0507 & \textbf{0.2049} & 0.1290 \\
    CLIP-DINOiser & \textbf{0.3555} & 0.3255 & 0.2807 \\
    Grounded-SAM & 0.5822 & 0.4714 & \textbf{0.6010} \\
    C-RADIOv3 & \textbf{0.2718} & 0.2644 & 0.0658 \\
    SAM3 & 0.6162 & 0.5777 & \textbf{0.6190} \\
    SigLIP & 0.2580 & 0.2336 & \textbf{0.3391} \\
    YOLOE-seg & 0.3785 & 0.3850 & \textbf{0.4754} \\
    \hline
\end{tabular}
\caption{Text-to-mask performance of conventional classes across multiple datasets.}
\label{tab:exp1_conclass_multidataset_F1}
\end{table}

\subsection{Experiment 2: Mask-to-Top-K Semantic Masks}
\label{subsec:results_exp2}

\begin{figure}[t]
    \centering
    \includegraphics[trim={0cm 0.3cm 0cm 0.0cm},clip,width=\columnwidth]{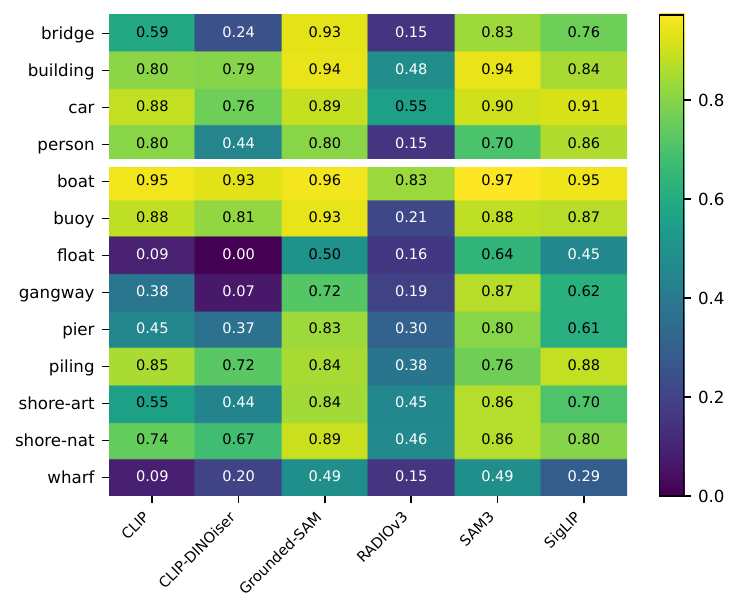}
    \caption{Mask-to-mask top-1 accuracy of each class in the Hawaii dataset.}
    \label{fig:exp2_perclass_heatmap_top1}
\end{figure}

Experiment 2 evaluates each model's ability to accurately match visual regions with other visual regions that exhibit similar characteristics across images.
Fig.~\ref{fig:exp2_perclass_heatmap_top1} shows the top-1 accuracy for each class and model in the Hawaii dataset, with landscape classes excluded to focus the comparison on conventional and coastal classes.
Classes are divided into two semantic groups, consistent with Fig.~\ref{fig:exp1_perclass_heatmap_F1}: conventional and coastal.

Fig.~\ref{fig:exp2_perclass_heatmap_top1} shows that models are generally able to accurately match visual regions across images, regardless of whether the regions belong to conventional or coastal semantic classes.
Although some models consistently outperform others, individual models generally achieve similar top-1 accuracy across the two semantic groups.
This suggests that the lower text-to-mask performance observed for coastal classes is not solely attributable to differences in visual representation.

\subsection{Experiment 3: Mask-to-Top-K Text Classes}
\label{subsec:results_exp3}

Experiment 3 evaluates each model's ability to accurately associate given visual regions with correct classifying text.
Table~\ref{tab:exp3_variable_prompts} reports top-1 accuracy for the evaluated models using different word banks.
Each section corresponds to a different trial within experiment 3, with the original class labels shown first and alternative labels highlighted in gray.
In the first trial, each ground-truth mask was associated with its original word label.
In the second trial, alternative words were mapped to the same ground-truth masks.
Specifically, \texttt{shore-artificial} and \texttt{shore-natural} were both mapped to \texttt{shore}.
Similarly, \texttt{pier} and \texttt{wharf} were both mapped to \texttt{dock}.
Lastly, \texttt{piling} was mapped to \texttt{pole}. 

In the \texttt{pier+wharf} and \texttt{piling} groups, remapping the classes to the more general or colloquial terms \texttt{dock} and \texttt{pole}, respectively, improves accuracy for all models.
This suggests that some of the observed difficulty in coastal classification may stem from limited familiarity with specific marine terminology.
The \texttt{shore-artificial+shore-natural} remapping to \texttt{shore} does not consistently improve performances across models, but it achieves higher accuracy than at least one of the original labels for every model.
Taken together, these results provide further evidence that linguistic representation can influence VLM performance on coastal visual features, suggesting that the choice of descriptive terminology may be an important factor in vision-language alignment.

\begin{table}[t]
\centering
\begin{tabular}{lllll}
    \hline
    class & CLIP & CLIP-DINOiser & C-RADIOv3 & SigLIP \\
    \hline
    shore-artificial & \textbf{0.3058} & 0.0744 & 0.0307 & \textbf{0.7190} \\
    shore-natural & 0.0000 & 0.7125 & \textbf{0.2308} & 0.0246 \\
    \rowcolor{gray!20}
    shore & 0.1857 & \textbf{0.8026} & 0.1714 & 0.4234 \\
    \hline
    pier & 0.0000 & 0.1098 & 0.0062 & 0.0061 \\
    wharf & 0.0857 & 0.0571 & 0.0294 & 0.0571 \\
    \rowcolor{gray!20}
    dock & \textbf{0.2814} & \textbf{0.7236} & \textbf{0.1590} & \textbf{0.1608} \\
    \hline
    piling & 0.7381 & 0.0095 & 0.0331 & 0.3095 \\
    \rowcolor{gray!20}
    pole & \textbf{0.7619} & \textbf{0.0762} & \textbf{0.1325} & \textbf{0.6238} \\
    \hline
\end{tabular}
\caption{Mask-to-text top-1 accuracy using variable word banks.}
\label{tab:exp3_variable_prompts}
\end{table}

\vspace{-0.1cm}
\section{Conclusion}
This work presents a systematic evaluation of seven VLMs in coastal environments using a densely labeled dataset collected across diverse regions of Oahu, Hawaii. 
Across three complementary experiments, we evaluate text-to-mask, mask-to-mask, and mask-to-text alignment to examine how VLMs associate coastal visual regions with semantic concepts. 
Overall, landscape classes achieve stronger text-to-mask performance than conventional and coastal classes, while mask-to-mask results show that models can generally identify visually similar regions across both groups.

Taken together, these results suggest that lower performance on coastal classes cannot be attributed solely to environmental context or visual representation. 
Instead, segmentation difficulty and linguistic representation appear to be important contributing factors. 
In particular, replacing specific marine terms with more general or colloquial alternatives improves recognition for several coastal classes, indicating that the choice of terminology can influence vision-language alignment.

Future work will extend this evaluation to additional coastal environments and datasets, investigate domain shift more systematically, and explore methods for improving VLM performance through better segmentation and domain-specific linguistic representations.
\balance

\vspace{0.5cm}
\bibliographystyle{IEEEtran}
\bibliography{ref}
\end{document}